\pdfoutput=1

\documentclass[11pt]{article}

\usepackage[preprint]{acl}

\usepackage{times}
\usepackage{latexsym}

\usepackage[T1]{fontenc}

\usepackage[utf8]{inputenc}

\usepackage{microtype}

\usepackage{inconsolata}

\usepackage{graphicx}
\usepackage{multirow}
\usepackage{pifont}
\usepackage{xcolor}
\usepackage{booktabs}
\usepackage{hyperref}
\usepackage{makecell}
\usepackage{amsmath}
\usepackage[tikz]{bclogo}
\usepackage{amsfonts}
\usepackage{enumitem}
\usepackage{stackengine}
\usepackage{multicol}
\usepackage[ruled,vlined]{algorithm2e}
\usepackage{algpseudocode}
\usepackage{soul}
\usepackage{colortbl}
\usepackage{enumitem}

\usepackage[most]{tcolorbox}
\tcbuselibrary{skins}
\tcbuselibrary{breakable}

\definecolor{my_blue}{RGB}{0,120,255}
\definecolor{my_purple}{RGB}{161, 27, 155}
\definecolor{my_green}{RGB}{0, 176, 80}
\definecolor{msftBlue}{RGB}{0,164,239}
\definecolor{msftGreen}{RGB}{127,186,0}
\definecolor{msftYello}{RGB}{255,185,0}
\definecolor{msftBlack}{RGB}{0,0,0}
\newcommand{\finding}[1]{
\begin{bclogo}[couleur= msftBlack!05, epBord= 1, arrondi=0.1, logo=\bclampe,marge= 2, ombre=true, blur, couleurBord=msftBlack!10, tailleOndu=3, sousTitre ={\em #1}]{} 
\end{bclogo}
}

\definecolor{my-blue}{RGB}{0, 69, 177}
\definecolor{my-purple}{RGB}{105,50,230}

\title{\textit{The Good, The Bad, and The Greedy:} \\ 
Evaluation of LLMs Should Not Ignore Non-Determinism}

\author{Yifan Song$^{\heartsuit}$\qquad Guoyin Wang\qquad Sujian Li$^{\heartsuit}$ \qquad Bill Yuchen Lin$^{\clubsuit}$\\[10pt]
$^\heartsuit$Peking University \quad \quad $^\clubsuit$Allen Institute for AI \\[5pt]
{{\texttt{yfsong@pku.edu.cn \quad yuchenl@allenai.org}}}
}

\begin{document}
\maketitle
\begin{abstract}

\renewcommand{\paragraph}[1]{\noindent \textbf{#1}}

Current evaluations of large language models (LLMs) often overlook non-determinism, typically focusing on a single output per example. This limits our understanding of LLM performance variability in real-world applications. 
Our study addresses this issue by exploring key questions about the performance differences between greedy decoding and sampling, identifying benchmarks’ consistency regarding non-determinism, and examining unique model behaviors. 
Through extensive experiments, we observe that greedy decoding generally outperforms sampling methods for most evaluated tasks.
We also observe consistent performance across different LLM sizes and alignment methods, noting that alignment can reduce sampling variance. 
Moreover, our best-of-N sampling approach demonstrates that smaller LLMs can match or surpass larger models such as GPT-4-Turbo, highlighting the untapped potential of smaller LLMs. 
This research shows the importance of considering non-determinism in LLM evaluations and provides insights for future LLM development and evaluation.
\footnote{Code and data are available at \href{https://github.com/Yifan-Song793/GoodBadGreedy}{https://github.com/Yifan-Song793/GoodBadGreedy}}
\end{abstract}

\section{Introduction}

When evaluating a large language model (LLM), two common generation configurations are commonly used: greedy decoding and nucleus sampling~\citep{holtzman2019curious}. It’s important to note that given a particular input, the same LLM may generate significantly different outputs under various decoding configurations, a phenomenon known as non-determinism in generation. 
However, most evaluations of LLMs are based on a single output per example. 
This practice is primarily due to practical considerations, as LLM inference and evaluation can be computationally expensive. 
Neglecting non-determinism in generation significantly limits our comprehensive understanding of LLMs. 
Additionally, without reporting the standard deviation in most current LLM evaluations, it is difficult to measure the variability and dynamics of LLMs in real-world applications.

For certain capabilities such as math reasoning~\citep{cobbe2021training,hendrycks2021measuring} and coding, greedy generation is preferred to ensure fair comparisons. Nonetheless, it remains unclear whether there are significant differences in performance between greedy decoding and sampling. Recent investigations have also highlighted potential issues of instability in LLMs~\citep{li2024common,hassid2024larger}. 
In a study where the best answer was selected from 256 random generations, the Llama-2-7B model achieved an impressive 97.7\% accuracy in solving GSM8K questions, even surpassing GPT-4~\citep{li2024common}. This phenomenon further underscores the enormous potential of LLMs in their non-deterministic outputs.

Herein, we aim to investigate a series of critical questions regarding the non-determinism of LLM generations, which have not been fully explored:

\begin{itemize}[nosep, leftmargin=*]
    \item \textbf{Q1}: \textit{How does the performance gap between greedy decoding and sampling differ?}
    \item \textbf{Q2}: \textit{When is greedy decoding better than sampling, and vice versa? Why?}
    \item \textbf{Q3}: \textit{Which benchmark is most/least consistent with respect to non-determinism?}
    \item \textbf{Q4}: \textit{Do any models possess unique patterns?}
\end{itemize}

\noindent
Apart from Q1-Q4 in Sec.~\ref{ssec:main}, we also explore the \textbf{\textit{scaling}} effect on non-determinism (Sec.~\ref{ssec:scaling}), the \textbf{\textit{alignment}} effect on non-determinism (Sec.~\ref{ssec:align}), the \textbf{\textit{temperature}} and \textbf{\textit{repetition}} effect on generation, and the full \textbf{\textit{potential}} of LLMs (Sec.~\ref{ssec:potential}).

Our extensive results reveal these findings:
\begin{itemize}[leftmargin=*, nolistsep]
\setlength{\itemsep}{1mm}
\item For most benchmarks we evaluated, a notable performance gap is observed between greedy generation and the average score of multiple sampling. In certain cases, the performance ranking under different generation configurations differs.
\item Greedy decoding exhibits superior performance than sampling methods on most evaluated benchmarks, except for AlpacaEval where sampling shows higher win rate.
\item LLMs displayed consistent performance across different generation configurations for benchmarks with constrained output spaces, such as MMLU and MixEval. Notably, tasks involving math reasoning and code generation were most impacted by sampling variance.
\item The above findings remain consistent across different sizes and families of LLMs.
\item Alignment methods, e.g., DPO~\citep{rafailov2024direct}, can significantly reduce the sampling variance for most benchmarks.
\item High temperature will significantly harm the reasoning and code generation capabilities of LLMs, while higher repetition penalty leads to improved performance on AlpacaEval.
\item In the best-of-N sampling setting, 7B-level LMs have the potential to outperform GPT-4-Turbo, and cutting-edge reward models can select superior responses from multiple sampled candidates.
\end{itemize}

\section{Experimental Setup}


\paragraph{Benchmarks.}

We select multiple benchmarks for our experiments, encompassing abilities of general instruction-following, knowledge, math reasoning, coding, etc.
As summarized in Table~\ref{tab:dataset}, the selected benchmarks are: AlpacaEval 2~\citep{alpaca_eval}, Arena-Hard~\citep{arenahard2024}, WildBench v2~\citep{lin2024wildbench}, MixEval~\citep{ni2024mixeval}, MMLU-Redux~\citep{gema2024we}, GSM8K~\citep{cobbe2021training}, and HumanEval~\citep{chen2021evaluating}.

AlpacaEval 2~\citep{alpaca_eval}, Arena-Hard~\citep{arenahard2024} and WildBench v2~\citep{lin2024wildbench} are general instruction-following benchmarks.
AlpacaEval consists of 805 questions, Arena-Hard incorporating 500 well-defined technical problem-solving queries, and WildBench including 1024 challenging tasks from real users.
For AlpacaEval 2, we report the length-controlled win rate (LC).
For Arena-Hard, we report the win rate (WR) against the baseline model.
For WildBench, we use task-wise scores and the corresponding task-macro WB-Score as the metrics.

Since the original MMLU~\citep{hendrycks2020measuring} benchmark is huge and contain numerous ground truth errors~\citep{wang2024mmlu,gema2024we}, we use MMLU-Redux~\citep{gema2024we} which is a subset of 3000 manually re-annotated questions across 30 MMLU subjects.
We also include GSM8K~\citep{cobbe2021training}, and HumanEval~\citep{chen2021evaluating}, two popular benchmarks for evaluating the math and code generation abilities of LLMs.

\begin{table}[t]
\centering
\resizebox{\linewidth}{!}{
\begin{tabular}{lccc}
\toprule
\textbf{Dataset} & \textbf{Instance Num.} & \textbf{Sample Num.} & \textbf{Metric} \\
\midrule
AlpacaEval 2 & 805 & 16 & LC \\
Arena-Hard & 500 & 16 & WR \\
MixEval & 4000 & 16 & Score \\
WildBench v2 & 1024 & 16 & WB-Score \\
MMLU-Redux & 3000 & 32 & Acc \\
GSM8K & 1319 & 128 & EM \\
HumanEval & 164 & 128 & Pass@1 \\
\bottomrule
\end{tabular}
}
\caption{
Statistics of datasets.
}
\label{tab:dataset}
\end{table}

\paragraph{LLMs.}
We test several open-weight LLMs, including Llama-3-Instruct~\citep{llama3}, Yi-1.5-Chat~\citep{young2024yi}, Qwen-2-Instruct~\citep{qwen}, Mistral~\citep{jiang2023mistral}, which are widely used.
A proprietary LLM, GPT-4-Turbo, is included for comparison.
We also consider models of different sizes in the same family such as Qwen2 and Yi-1.5  for more analysis.
To study the effect of alignment techniques, we evaluate models trained with different alignment methods, including DPO~\citep{rafailov2024direct}, KTO~\citep{ethayarajh2024kto}, IPO~\citep{azar2024general}, ORPO~\citep{hong2024reference}, RDPO~\citep{park2024disentangling}, and SimPO~\citep{meng2024simpo}.
We use the checkpoints released by \citet{meng2024simpo}.

\paragraph{Setup.}
We aim to compare the performance of LLMs under different decoding configurations.
We select greedy decoding and sampling generation for the main comparison.
For sampling, we set the temperature to $1.0$ and top-p to $1.0$.

We use official evaluation scripts for AlpacaEval 2, Arena-Hard, WildBench, and MixEval.
For MMLU-Redux, instead of using the next token probability of the choice letters, we employ zero-shot CoT and encourage the model to generate the answer in the form of natural language sentence.
For GSM8K and HumanEval, we use Open-Instruct framework~\citep{wang2023far} to evaluate the models, which may differ from zero-shot CoT. We will run more comprehensive evaluations on these two benchmarks in the future.
We sample 16 completions for AlpacaEval 2, Arena-Hard, WildBench, and MixEval, 32 completions for MMLU-Redux, 128 for GSM8K and HumanEval.

\begin{table*}[t]
\centering
\resizebox{\linewidth}{!}{
\begin{tabular}{c|cccc|cccc|cccc}
\toprule
\multirow{2}{*}{\textbf{Model}} & \multicolumn{4}{c|}{\textbf{AlpacaEval 2 (N=16)}} & \multicolumn{4}{c|}{\textbf{Arena-Hard (N=16)}} & \multicolumn{4}{c}{\textbf{MixEval (N=16)}} \\
\cmidrule(l){2-5} \cmidrule(l){6-9} \cmidrule(l){10-13} 
& \textbf{Greedy} & \textbf{Sample} & \textbf{Std.} & \textbf{$\Delta$} & \textbf{Greedy} & \textbf{Sample} & \textbf{Std.} & \textbf{$\Delta$} & \textbf{Greedy} & \textbf{Sample} & \textbf{Std.} & \textbf{$\Delta$} \\
\midrule
GPT-4-Turbo & \cellcolor{red!2.5}49.6 & \cellcolor{red!2.5}50.1 & 0.76 & 2.5 & \cellcolor{green!24.5}80.1 & \cellcolor{green!24.5}75.2 & \textbf{1.31} & 3.6 & \cellcolor{green!2}89.2 & \cellcolor{green!2}88.8 & 0.18 & 0.8 \\
Llama-3-8B-Instruct & \cellcolor{red!12}26.8 & \cellcolor{red!12}29.2 & 0.88 & 2.8 & \cellcolor{green!25.5}23.5 & \cellcolor{green!25.5}18.4 & 0.71 & 2.7 & \cellcolor{green!10.5}74.6 & \cellcolor{green!10.5}72.5 & 0.25 & 0.9 \\
Yi-1.5-6B-Chat & \cellcolor{red!2.5}17.5 & \cellcolor{red!2.5}18.0 & 0.91 & 3.4 & \cellcolor{green!9.5}13.7 & \cellcolor{green!9.5}11.8 & 0.88 & 3.1 & \cellcolor{green!7}70.0 & \cellcolor{green!7}68.6 & 0.26 & 1.0 \\
Yi-1.5-9B-Chat & \cellcolor{red!5}23.1 & \cellcolor{red!5}24.1 & 0.91 & 3.4 & \cellcolor{green!29}32.8 & \cellcolor{green!29}27.0 & 1.25 & 4.4 & \cellcolor{green!6.5}74.0 & \cellcolor{green!6.5}72.7 & \textbf{0.35} & 1.4 \\
Yi-1.5-34B-Chat & \cellcolor{red!0.5}34.9 & \cellcolor{red!0.5}35.0 & 0.99 & 3.9 & \cellcolor{green!9.5}42.8 & \cellcolor{green!9.5}40.9 & 1.82 & 5.7 & \cellcolor{green!0.5}81.9 & \cellcolor{green!0.5}81.8 & 0.47 & 1.5 \\
Qwen2-7B-Instruct & \cellcolor{red!4.5}18.2 & \cellcolor{red!4.5}19.1 & \textbf{2.51} & 8.6 & \cellcolor{green!38}23.7 & \cellcolor{green!38}16.1 & 0.87 & 3.1 & 76.2 & 76.2 & 0.21 & 0.6 \\
Mistral-7B-Instruct-v0.2 & \cellcolor{green!12}15.4 & \cellcolor{green!12}13.0 & 1.02 & 4.2 & \cellcolor{red!0.5}12.5 & \cellcolor{red!0.5}12.6 & 0.57 & 2.0 & \cellcolor{red!1}69.8 & \cellcolor{red!1}70.0 & 0.24 & 0.9 \\
\toprule
\multirow{2}{*}{\textbf{Model}} & \multicolumn{4}{c|}{\textbf{MMLU-Redux (N=32)}} & \multicolumn{4}{c|}{\textbf{GSM8K (N=128)}} & \multicolumn{4}{c}{\textbf{HumanEval (N=128)}} \\
\cmidrule(l){2-5} \cmidrule(l){6-9} \cmidrule(l){10-13} 
& \textbf{Greedy} & \textbf{Sample} & \textbf{Std.} & \textbf{$\Delta$} & \textbf{Greedy} & \textbf{Sample} & \textbf{Std.} & \textbf{$\Delta$} & \textbf{Greedy} & \textbf{Sample} & \textbf{Std.} & \textbf{$\Delta$} \\
\midrule
GPT-4-Turbo & \cellcolor{green!1}82.6 & \cellcolor{green!1}82.4 & 0.43 & 1.6 & \cellcolor{green!3.5}84.5 & \cellcolor{green!3.5}83.8 & 0.77 & 2.5 & \cellcolor{green!16.5}89.6 & \cellcolor{green!16.5}84.1 & 2.65 & 11.0 \\
Llama-3-8B-Instruct & \cellcolor{red!14.5}47.8 & \cellcolor{red!14.5}50.7 & \textbf{0.70} & 2.8 & \cellcolor{green!7}67.6 & \cellcolor{green!7}64.4 & \textbf{2.50} & 13.4 & \cellcolor{green!80}58.5 & \cellcolor{green!80}31.8 & 3.62 & 18.3 \\
Yi-1.5-6B-Chat & \cellcolor{green!12.5}52.1 & \cellcolor{green!12.5}49.6 & 0.67 & 2.5 & \cellcolor{green!7}74.5 & \cellcolor{green!7}73.1 & 0.92 & 4.1 & \cellcolor{green!37.5}48.2 & \cellcolor{green!37.5}35.7 & 4.86 & 19.5 \\
Yi-1.5-9B-Chat & \cellcolor{green!6}65.5 & \cellcolor{green!6}64.3 & 0.53 & 2.3 & \cellcolor{green!9.5}82.9 & \cellcolor{green!9.5}81.0 & 0.69 & 3.9 & \cellcolor{green!57.3}55.5 & \cellcolor{green!57.3}36.4 & \textbf{4.92} & 27.5 \\
Yi-1.5-34B-Chat &  \cellcolor{green!5}83.2 & \cellcolor{green!5}82.2 & 0.34 & 1.1 & \cellcolor{green!18.5}85.4 & \cellcolor{green!18.5}81.7 & 0.56 & 2.9 & \cellcolor{green!45.9}64.6 & \cellcolor{green!45.9}49.3 & 4.08 & 21.4 \\
Qwen2-7B-Instruct & \cellcolor{green!13.5}64.4 & \cellcolor{green!13.5}61.7 & 0.46 & 2.1 & \cellcolor{green!57.5}83.5 & \cellcolor{green!57.5}72.0 & 1.74 & 11.3 & \cellcolor{green!58.5}67.7 & \cellcolor{green!58.5}48.2 & 4.68 & 27.4 \\
Mistral-7B-Instruct-v0.2 & \cellcolor{green!6.5}49.7 & \cellcolor{green!6.5}48.4 & 0.49 & 2.2 & \cellcolor{green!19.5}45.9 & \cellcolor{green!19.5}42.0 & 0.99 & 5.1 & \cellcolor{green!35.7}37.8 & \cellcolor{green!35.7}25.9 & 2.52 & 14.0 \\
\bottomrule
\end{tabular}
}
\caption{
Results on six popular benchmarks. ``Sample'' and ``Std.'' denotes the average score and the standard deviation of ``N'' runs under sampling setup. ``$\Delta$'' denotes the performance gap between the best and worst run. Scores where greedy decoding surpasses the sampling average are highlighted in \textcolor{my_green}{green}, while those lower are marked in \textcolor{red}{red}. The intensity of the color indicates the magnitude of the difference (best viewed in color).
}
\label{tab:main}
\end{table*}

\begin{table*}[t]
\centering
\resizebox{\linewidth}{!}{
\begin{tabular}{l|cccc|cccc|cccc}
\toprule
\multirow{2}{*}{\textbf{Metric}} & \multicolumn{4}{c|}{\textbf{Llama-3-8B-Instruct}} & \multicolumn{4}{c|}{\textbf{Yi-1.5-6B-Chat}} & \multicolumn{4}{c}{\textbf{Qwen2-7B-Instruct}} \\
\cmidrule(l){2-5} \cmidrule(l){6-9} \cmidrule(l){10-13} 
& \textbf{Greedy} & \textbf{Sample} & \textbf{Std.} & \textbf{$\Delta$} & \textbf{Greedy} & \textbf{Sample} & \textbf{Std.} & \textbf{$\Delta$} & \textbf{Greedy} & \textbf{Sample} & \textbf{Std.} & \textbf{$\Delta$} \\
\midrule
WB-Score & \cellcolor{green!17}29.6 & \cellcolor{green!17}26.2 & 1.65 & 5.7 & \cellcolor{green!7.5}23.9 & \cellcolor{green!7.5}22.4 & 1.67 & 5.3 & \cellcolor{green!44.5}32.7 & \cellcolor{green!44.5}23.8 & 2.13 & 7.7 \\
\midrule
Creative Tasks & \cellcolor{red!1}42.2 & \cellcolor{red!1}42.4 & 1.77 & 6.7 & 32.1 & 32.1 & 2.33 & 10.3 & \cellcolor{green!41}39.6 & \cellcolor{green!41}31.4 & 2.21 & 8.5 \\
Planning \& Reasoning & \cellcolor{green!12}33.8 & \cellcolor{green!12}31.4 & 1.19 & 3.6 & \cellcolor{green!2.5}27.9 & \cellcolor{green!2.5}27.4 & 1.77 & 5.7 & \cellcolor{green!39.5}36.0 & \cellcolor{green!39.5}28.1 & 1.95 & 6.1 \\
Math \& Data Analysis & \cellcolor{green!9}17.8 & \cellcolor{green!9}16.0 & 2.85 & 9.2 & \cellcolor{red!1}17.4 & \cellcolor{red!1}17.5 & 1.99 & 6.5 & \cellcolor{green!45.5}27.6 & \cellcolor{green!45.5}18.5 & 2.69 & 10.4 \\
Info/Advice Seeking & \cellcolor{green!8}39.0 & \cellcolor{green!8}37.4 & 1.30 & 5.5 & \cellcolor{green!11.5}32.5 & \cellcolor{green!11.5}30.2 & 1.80 & 6.3 & \cellcolor{green!40.5}40.3 & \cellcolor{green!40.5}32.2 & 1.84 & 6.5 \\
Coding \& Debugging & \cellcolor{green!40.5}24.1 & \cellcolor{green!40.5}16.0 & 3.12 & 10.9 & \cellcolor{green!19.5}16.7 & \cellcolor{green!19.5}12.8 & 1.70 & 5.4 & \cellcolor{green!54}26.3 & \cellcolor{green!54}15.5 & 2.82 & 9.3 \\
\bottomrule
\end{tabular}
}
\caption{
Results on \textbf{WildBench v2}, with sampling N=16 generations for each model. In addition to WB-Score, we also report the score for each task category.
}
\label{tab:wb}
\end{table*}

\section{Experimental Results}
\label{ssec:main}


We present our experiment results in Table~\ref{tab:main} and Table~\ref{tab:wb}.
We analyze the results and answer several important research questions around the non-determinism of LLM generations as follows.

\finding{
Q1. How does the performance gap between greedy decoding and sampling differ?
}
From the results, we observe a consistent performance gap between greedy decoding and the sampling method.
This disparity holds true across various LLMs, 
whether they are proprietary or open-source, 
and across multiple benchmarks encompassing instruction-following, 
language understanding, math reasoning, and code generation.
For WildBench, which enables fine-grained analysis of LLM capabilities, the performance gap is also evident across all task categories, as shown in Table~\ref{tab:wb}.
Different decoding configurations can even alter the model rankings in some cases.
For example, on Arena-Hard, Qwen2-7B is slightly better than Llama-3-8B when both use greedy decoding; However, Llama-3-8B may outperform Qwen2-7B when both decode by sampling.

\finding{
Q2. When is greedy decoding better than sampling, and vice versa? Why?
}

For most evaluated tasks and models, greedy decoding outperforms sampling.
However, AlpacaEval serves as a notable exception, where sampling demonstrates superior performance.

GSM8K and HumanEval are reasoning tasks requiring LLMs to solve specific math or coding problems with definite solutions. 
MixEval also follows a deterministic pattern with its ground-truth-based benchmarks. 
While AlpacaEval, Arena-Hard, and WildBench are open-ended instruction-following benchmarks, AlpacaEval exhibits a contrasting behavior compared to the others.
The potential reasons are two folds:
Firstly, the task category distributions vary across different benchmarks. 
As highlighted by \citet{lin2024wildbench}, 50\% of instances in AlpacaEval are information-seeking, whereas more than 50\% in Arena-Hard are related to coding and debugging.
Furthermore, the difficulty of instances might play an important role. 
The tasks in both Arena-Hard and WildBench, sourced from real users, pose substantial challenges. On the other hand, instances in AlpacaEval are comparatively simpler.

In summary:
1) Greedy decoding generally proves more effective for most tasks.
2) In the case of AlpacaEval, which comprises relatively simpler open-ended creative tasks, sampling tends to generate better responses.

\finding{
Q3. Which benchmark is most/least consistent with respect to non-determinism?
}

\noindent
MixEval and MMLU exhibit the highest stability, either in terms of the performance gap between greedy decoding and sampling or the standard deviation across different samplings. This stability can be attributed to the constrained answer space of these benchmarks. Specifically, MMLU is structured in a multiple-choice format, and MixEval, comprising various ground-truth-based benchmarks, prompts LLMs to generate short answers, further limiting the output space.

In contrast, GSM8K and HumanEval are relatively less stable with respect to non-deterministic generations. The performance gap between the best and worst samplings can exceed 10.0 points. 

\finding{
Q4. Do the models possess distinctive characteristics?
}

GPT-4-Turbo shows consistent performance across multiple tasks, with a smaller performance gap between greedy decoding and sampling, as well as improved sampling quality. Some open-weight LLMs, however, exhibit unique characteristics. For example, Mistral-7B-Instruct-v0.2 displays inverse behavior on open-ended instruction following tasks like AlpacaEval and Arena-Hard when compared to other models. Similarly, Llama-3-8B-Instruct performs better by sampling than by greedy decoding on MMLU, which is unlike the behavior of other models.

These observations raise intriguing questions for future research. Why do certain models exhibit inverse behavior on specific tasks? Can these unique characteristics be leveraged to develop more robust LLMs? These questions highlight the need for deeper explorations into the underlying mechanisms of LLMs. Such research could significantly enhance our understanding of how different models and training  impact model behavior.

\section{How Various Factors Influence Non-Determinism?}

In this section, we further investigate how various factors, such as scaling, alignment, and several decoding parameters, influence non-determinism.

\subsection{Scaling Effect on Non-Determinism} 
\label{ssec:scaling}

\begin{table}[t]
\centering
\resizebox{\linewidth}{!}{
\begin{tabular}{lccc|ccc}
\toprule
\multirow{2}{*}{\textbf{Model}} & \multicolumn{3}{c}{\textbf{AlpacaEval}} & \multicolumn{3}{c}{\textbf{MMLU}} \\
\cmidrule(l){2-4} \cmidrule(l){5-7}
& G & S & Std. & G & S & Std. \\
\midrule
Qwen2-0.5B-Instruct & \cellcolor{red!3}1.1 & \cellcolor{red!3}1.7 & 0.77 & \cellcolor{red!3}36.4 & \cellcolor{red!3}37.0 & 0.70 \\
Qwen2-1.5B-Instruct & \cellcolor{red!7}1.9 & \cellcolor{red!7}3.3 & 0.88 & \cellcolor{green!2.5}42.6 & \cellcolor{green!2.5}42.1 & 0.68\\
Qwen2-7B-Instruct & \cellcolor{red!4.5}18.2 & \cellcolor{red!4.5}19.1 & 2.51 & \cellcolor{red!3.5}61.0 & \cellcolor{red!3.5}61.7 & 0.46 \\
\toprule
\multirow{2}{*}{\textbf{Model}} & \multicolumn{3}{c}{\textbf{GSM8K}} & \multicolumn{3}{c}{\textbf{HumanEval}} \\
\cmidrule(l){2-4} \cmidrule(l){5-7}
& G & S & Std. & G & S & Std. \\
\midrule
Qwen2-0.5B-Instruct & \cellcolor{green!20}31.7 & \cellcolor{green!20}14.3 & 1.86 & \cellcolor{green!11}28.0 & \cellcolor{green!11}10.8 & 2.14 \\
Qwen2-1.5B-Instruct & \cellcolor{green!53}63.1 & \cellcolor{green!53}36.5 & 3.20 & \cellcolor{green!35}40.9 & \cellcolor{green!35}22.6 & 2.94 \\
Qwen2-7B-Instruct & \cellcolor{green!13}83.5 & \cellcolor{green!13}72.0 & 1.74 & \cellcolor{green!58}67.7 & \cellcolor{green!58}48.2 & 4.68 \\
\bottomrule
\end{tabular}
}
\caption{
Evaluation results on Qwen2-Instruct with different model sizes.
}
\label{tab:scale}
\end{table}



\begin{figure*}[t]
    \centering
    \includegraphics[width=\linewidth]{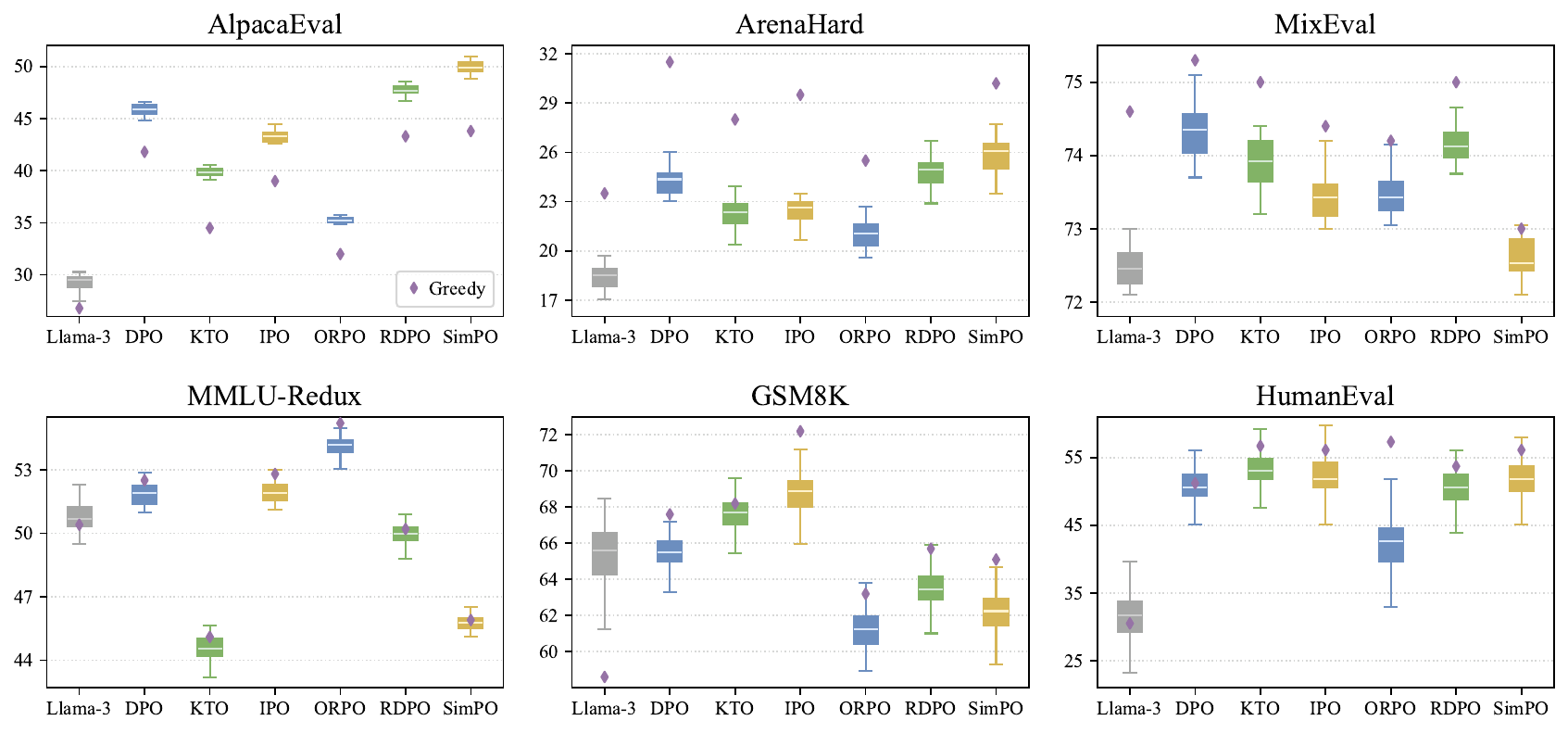}
    \caption{
    Alignment effects on non-determinism.
    }
    \label{fig:align}
\end{figure*}

\begin{figure}[t]
    \centering
    \includegraphics[width=\linewidth]{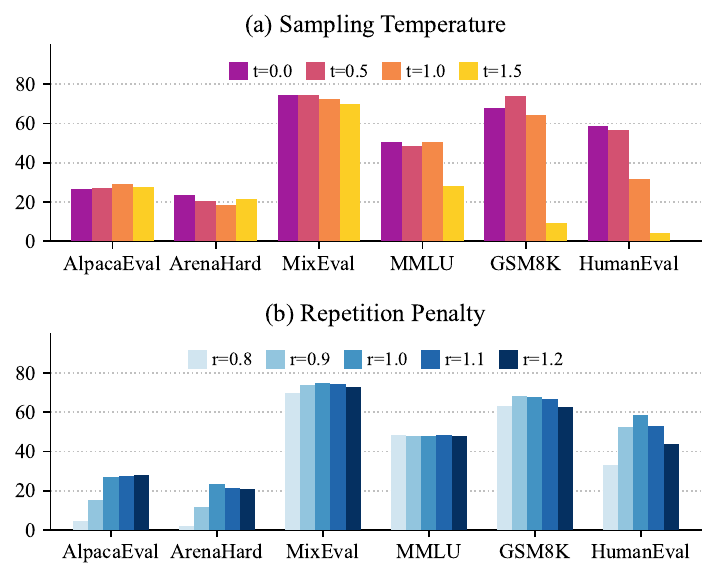}
    \caption{
    (a) Temperature effects on non-determinism. (b) Repetition penalty effects on generation. We compare performance of Llama-3-8B-Instruct with different generation parameters.
    }
    \label{fig:hyper}
\end{figure}

Some might assume that larger LMs will have lower uncertainty in decoding, leading to lower variance in performance when sampling. However, our results challenge this assumption. 

We use the Yi-1.5-Chat and Qwen2-Instruct series to investigate the scaling effect. The results for the Yi-1.5 and Qwen2 series are presented in Table~\ref{tab:main} and Table~\ref{tab:scale}, respectively. Performance differences are observed across LLMs of various sizes, ranging from 0.5B to 34B parameters. The findings in Section~\ref{ssec:main} are consistent across different model sizes. 
However, no pattern related to the number of model parameters could be identified. For instance, scaling parameters does not result in lower sampling variance. Notably, Qwen2-7B-Instruct shows higher variance on AlpacaEval and HumanEval compared to its smaller counterparts.

\subsection{Alignment Effect on Non-Determinism}
\label{ssec:align}

Alignment methods, such as DPO, enhance LLMs by learning from preference data. We evaluate the effects of alignment methods such as DPO, KTO, and SimPO, using Llama-3-8B-Instruct as the training starting point~\citep{meng2024simpo}. 

As shown in Figure~\ref{fig:align}, after applying these methods, both greedy decoding and sampling performances are affected. In several tasks, including AlpacaEval, MMLU, GSM8K, and HumanEval, a decrease in standard deviation is observed, suggesting that alignment may reduce the diversity of sampling outputs. However, it is crucial to note that not all alignment methods consistently improve model performance. For instance, KTO and SimPO lead to a performance decline in MMLU. Furthermore, SimPO's effectiveness appears limited on the recently introduced MixEval benchmark.

\subsection{Temperature Effect on Non-Determinism}
\label{ssec:temperature}


For sampling generation, temperature serves as a control mechanism for the randomness of the sampling process, where lower values make the model more deterministic, whereas higher values make the model more random.
In this section, we present an ablation study to evaluate the effect of varying temperatures on non-determinism generation.

As depicted in Figure~\ref{fig:hyper}(a), we observe that, for AlpacaEval, higher temperature will lead to slightly better performance, which aligns with the results in Sec.~\ref{ssec:main}.
A recent study~\citep{renze2024effect} finds that, on multiple-choice QA tasks, changes in temperature from 0.0 to 1.0 do not have a statistically significant impact on LLM performance.
Our results on MMLU aligns with their findings.
Another findings emerges when the temperature is extremely high, such as 1.5. 
Comparing with open-ended instruction following, a high temperature significantly impacts the reasoning and code generation capabilities of LLMs and the model struggles to solve questions in GSM8K and HumanEval.
However, it still manages to perform relatively well in open-ended instruction following tasks, such as AlpacaEval and ArenaHard.

\begin{figure*}[t]
    \centering
    \includegraphics[width=\linewidth]{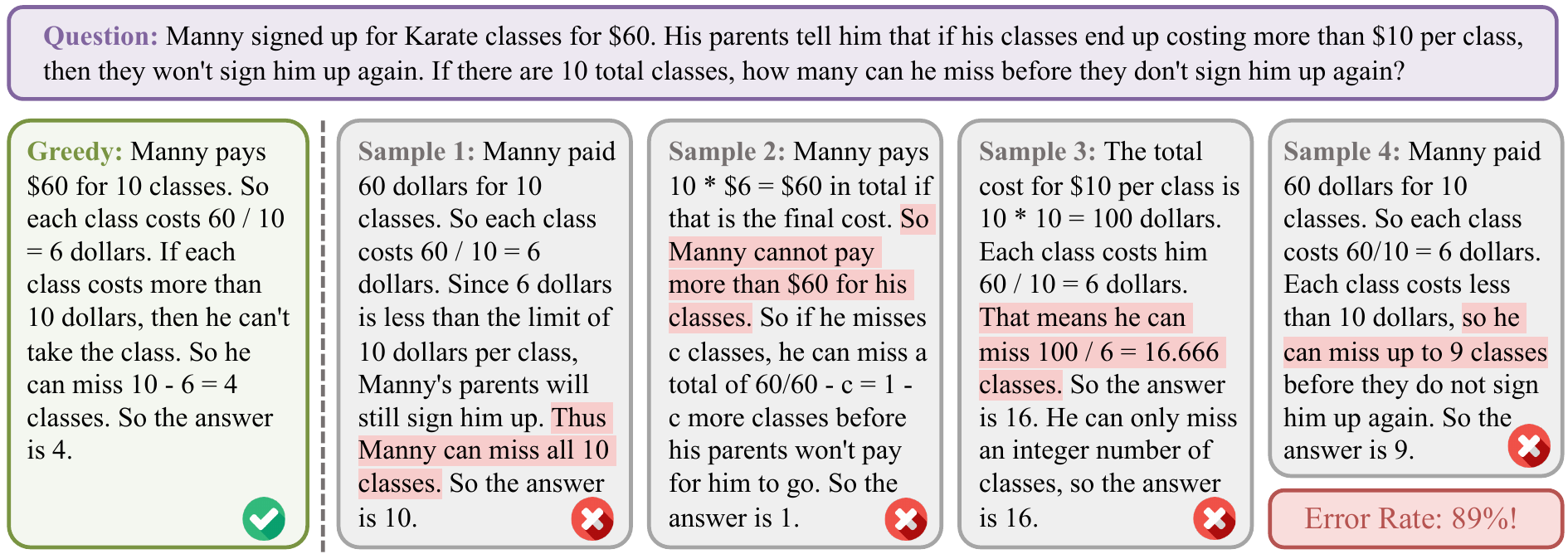}
    \caption{
    Case Study on non-determinism: Qwen2-7B-Instruct on GSM8K. Greedy decoding can effectively address the question. However, in 128 sampling generations for the same question, the error rate is 89\%.
    }
    \label{fig:case}
\end{figure*}
 
\subsection{Repetition Effect on Generation}
\label{ssec:repetition}


In addition to parameters that control greedy search and sampling, there are other parameters that influence the generation process, such as the repetition penalty~\citep{keskar2019ctrl}.
Here we examine the effect of repetition penalty on generation.
Repetition penalty penalizes new tokens based on whether they appear in the prompt and the generated text so far.
Values over 1.0 encourage the model to use new tokens, while values under 1.0 promote the reuse of tokens.
The default repetition penalty in generation is set at 1.0.

As illustrated in Figure~\ref{fig:hyper}(b), in most cases, it is advisable not to adjust this parameter, as maintaining the default value tends to yield the best performance.
For AlpacaEval, a higher repetition penalty like 1.2 results in marginally improved performance.
This improvement may be linked to GPT judges' preference for shorter, more concise answers.
Regarding MixEval and MMLU, repetition penalty has a minimal impact on the model's performance, since both benchmarks advocate for the model to generate concise answers.
Interestingly, for GSM8K, the model achieves the best performance when the repetition penalty is set at 0.9, and increasing this penalty parameter will cause a performance decline.
This phenomenon can be attributed to the nature of mathematical reasoning, which frequently necessitates the repetition of numbers and conditions outlined in the question.

\subsection{Surface Patterns in Non-Determinism Generation?}
\label{ssec:length}

\begin{table}[t]
\centering
\resizebox{\linewidth}{!}{
\begin{tabular}{l|cc|cc}
\toprule
\multirow{2}{*}{\textbf{Model}} & \multicolumn{2}{c|}{\textbf{AlpacaEval}} & \multicolumn{2}{c}{\textbf{ArenaHard}} \\
\cmidrule(l){2-3} \cmidrule(l){4-5}
& Len-G & Len-S & Len-G & Len-S \\
\midrule
GPT-4-Turbo & \cellcolor{my_blue!12}377 & \cellcolor{my_blue!12}389 & \cellcolor{my_blue!12}629 & \cellcolor{my_blue!12}641 \\
Llama-3-8B-Instruct & \cellcolor{my_blue!18}417 & \cellcolor{my_blue!18}435 & \cellcolor{my_purple!19}589 & \cellcolor{my_purple!19}570 \\
Yi-1.5-6B-Chat & \cellcolor{my_blue!2}477 & \cellcolor{my_blue!2}479 & \cellcolor{my_purple!34}670 & \cellcolor{my_purple!34}636 \\
Yi-1.5-9B-Chat & \cellcolor{my_blue!2}500 & \cellcolor{my_blue!2}502 & \cellcolor{my_blue!20}672 & \cellcolor{my_blue!20}692 \\
Yi-1.5-34B-Chat & \cellcolor{my_blue!3}450 & \cellcolor{my_blue!3}453 & \cellcolor{my_blue!12}693 & \cellcolor{my_blue!12}705 \\
Qwen2-7B-Instruct & \cellcolor{my_purple!10}420 & \cellcolor{my_purple!10}410 & \cellcolor{my_blue!21}573 & \cellcolor{my_blue!21}594 \\
Mistral-7B-Instruct-v0.2 & \cellcolor{my_blue!40}323 & \cellcolor{my_blue!40}372 & \cellcolor{my_blue!17}533 & \cellcolor{my_blue!17}550 \\
\toprule
\multirow{2}{*}{\textbf{Model}} & \multicolumn{2}{c|}{\textbf{MMLU}} & \multicolumn{2}{c}{\textbf{GSM8K}} \\
\cmidrule(l){2-3} \cmidrule(l){4-5}
& Len-G & Len-S & Len-G & Len-S \\
\midrule
GPT-4-Turbo & \cellcolor{my_blue!15}257 & \cellcolor{my_blue!15}272 & \cellcolor{my_blue!1}149 & \cellcolor{my_blue!1}150 \\
Llama-3-8B-Instruct & \cellcolor{my_purple!2}130 & \cellcolor{my_purple!2}128 & \cellcolor{my_blue!29}65 & \cellcolor{my_blue!29}94 \\
Yi-1.5-6B-Chat & \cellcolor{my_blue!13}145 & \cellcolor{my_blue!13}158 & \cellcolor{my_blue!5}127 & \cellcolor{my_blue!5}132 \\
Yi-1.5-9B-Chat & \cellcolor{my_blue!12}160 & \cellcolor{my_blue!12}172 & \cellcolor{my_blue!2}138 & \cellcolor{my_blue!2}140 \\
Yi-1.5-34B-Chat & \cellcolor{my_blue!9}263 & \cellcolor{my_blue!9}272 & \cellcolor{my_purple!1}143 & \cellcolor{my_purple!1}142 \\
Qwen2-7B-Instruct & \cellcolor{my_blue!15}75 & \cellcolor{my_blue!15}90 & \cellcolor{my_blue!18}121 & \cellcolor{my_blue!18}139 \\
Mistral-7B-Instruct-v0.2 & \cellcolor{my_blue!9}135 & \cellcolor{my_blue!9}144 & \cellcolor{my_blue!14}121 & \cellcolor{my_blue!14}135 \\
\bottomrule
\end{tabular}
}
\caption{
Length comparison. Cases where greedy decoding generates shorter responses than sampling average are highlighted in \textcolor{my_blue}{blue}, and marked in \textcolor{my_purple}{purple} vice versa.
}
\label{tab:length}
\end{table}

We try to explore the surface patterns in non-determinism generation.
Firstly, we compare the generation length of different generation configurations in Table~\ref{tab:length}.
The generation length for AlpacaEval and ArenaHard is defined as the length of the model's response, while for MMLU and GSM8K, it refers to the length of the final answer with chain-of-thoughts.
We observe that the completions generated by greedy decoding are typically marginally shorter than those produced via sampling generation.
However, this pattern deviates in the case of Yi series models on AlpacaEval and GSM8K, where the lengths of responses produced by both greedy decoding and sampling methods are comparable.

We also take Qwen2-7B-Instruct on GSM8K as a case study, where the greedy decoding significantly outperforms the sampling generation (83.5 vs. 72.0).
As depicted in Figure~\ref{fig:case}, greedy decoding solves the question effectively.
Nonetheless, when it is the turn for sampling generation, the error rate surges to 89\% within 128 responses.
This observation suggests that the sampling method could potentially harm reasoning capabilities for LLMs.

\begin{figure*}[t]
    \centering
    \includegraphics[width=\linewidth]{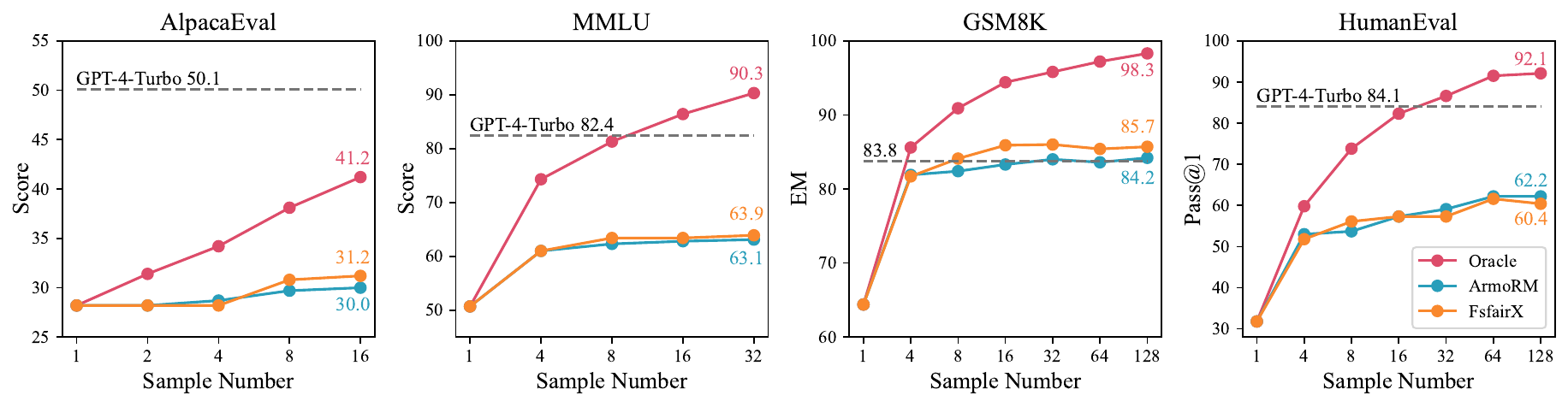}
    \caption{
    Potential of Llama-3-8B-Instruct. We use the setting of ``Best-of-N'', which selects the best response from N outputs for each example. We employ off-the-shelf reward models to rank the responses and select the one with the highest reward, while ``Oracle'' means the upper bound of Best-of-N method.
    }
    \label{fig:potential}
\end{figure*}

\section{What is the Full Potential of Non-Determinism?}
\label{ssec:potential}


Current evaluations of LLMs mainly assess them based on a single output per instance, which limits our understanding of their full potential. 
Following \citet{Jiang2023LLMBlenderEL} and \citet{li2024common}, we adopt a Best-of-N setting, selecting the best answer from $N$ sampled responses. 
To accomplish this, we employ off-the-shelf reward models, such as ArmoRM~\citep{ArmoRM} and FsfairX~\citep{xiong2024iterative}, to rank the responses of Llama-3-8B-Instruct, selecting the one with the highest reward.
We also include an ``oracle'' baseline which directly picks the best response as the upper bound of best-of-N strategy.

The results are depicted in Figure~\ref{fig:potential}. 
We observe a significant performance enhancement when applying simple best-of-N strategy for multiple sampled responses.
Notably, with the oracle selection, \textbf{even smaller LLMs like Llama-3-8B-Instruct can outperform GPT-4-Turbo on MMLU, GSM8K, and HumanEval}.
This finding underscores that compact-sized LLMs already exhibit robust capabilities, highlighting that a more significant challenge in alignment is to robustly decode such knowledge and reasoning paths.
Furthermore, cutting-edge reward models can also select superior responses from multiple generations, and can outperform GPT-4-Turbo on GSM8K with only 8 samples.
However, there is still a huge performance gap between reward models and the oracle baseline, indicating ample room for improvement.


Building upon these promising findings, there are two ways to further enhance the performance of smaller LLMs.
Firstly, probability calibration techniques can guide LLMs towards generating superior answers with higher likelihoods. Alignment methods, specifically preference optimization~\citep{rafailov2024direct}, play a pivotal role in this process. 
Secondly, strategies for ensemble learning or selecting the best answer from multiple completions warrant attention. 
Self-consistency~\citep{wang2022self} and advanced prompting techniques~\citep{Yao2023TreeOT,Lin2023SwiftSageAG}, which employs heuristic selection from multiple completions, is also worth further exploration.

\section{Related Work}

\paragraph{LLM Evaluation}

In recent years, the development of various benchmarks has significantly advanced the evaluation of LLMs.
Benchmarks like MMLU~\citep{hendrycks2020measuring}, HellaSwag~\citep{zellers2019hellaswag}, and ARC~\citep{clark2018think} have expanded the scope by assessing capabilities across knowledge understanding, and complex reasoning.
AlpacaEval~\citep{alpaca_eval}, MT-Bench~\citep{zheng2023judging}, ArenaHard~\citep{arenahard2024}, and WildBench~\citep{lin2024wildbench}, leveraging frontier models as judges, evaluate open-ended instruction-following capabilities.
Moreover, GSM8K~\citep{cobbe2021training}, MATH~\citep{hendrycks2021measuring}, HumanEval~\citep{chen2021evaluating}, and MBPP~\citep{austin2021program} focus on evaluating math reasoning and code generation capabilities.

Due to the costly nature of LLM inference and evaluation process, most evaluations of LLMs rely on a single output per example.
In this paper, we aim to explore the impact of various generation configurations, particularly non-deterministic generations, on the performance of LLMs.

\paragraph{Decoding Strategy}

Given a prompt, LLMs rely on a decoding strategy to auto-regressively generate response.
The simplest decoding method, greedy decoding, selects the next token with the highest probability.
Beam search~\citep{freitag2017beam}, an improved version of greedy search, retains the top-B tokens with the highest probability at each time step.
In order to generate diverse responses, non-determinism generation methods, such as Top-\textit{k}~\citep{fan2018hierarchical} and Top-\textit{p} sampling~\citep{holtzman2019curious}, randomly picks the next token based on the probability distribution.
The temperature parameter serves to balance response quality and diversity~\citep{ackley1985learning}.
Other decoding parameters, like length and repetition penalties~\citep{keskar2019ctrl}, are also available to further control the generation process.

\section{Conclusion \& Future directions}
We investigate a series of critical yet overlooked questions around non-determinism of LLM generations.
After evaluating several LLMs across seven commonly used benchmarks, we have answered several intriguing research questions.
Further analysis also provides insights on how scaling and alignment will effect on non-determinism generation.
We hope this work can enhance our comprehension of the generation methods and the widely used benchmarks.
Our evaluation results can also be used for improving future research. For example, our best-of-N results can serve as a benchmark for assessing reward models~\citep{Lambert2024RewardBenchER}.





\bibliography{custom}

\appendix

\end{document}